\documentclass[11pt]{article}
\pdfoutput=1
\usepackage[OT1]{fontenc}
\usepackage{fullpage}
\usepackage{microtype,enumitem}
\usepackage[round,authoryear]{natbib}
\usepackage{smile}
\usepackage[table]{xcolor}
\usepackage{colortbl}
\definecolor{LightCyan}{rgb}{0.8, 0.9, 1}
\newcolumntype{g}{>{\columncolor{LightCyan}}c}
\usepackage[colorlinks,linkcolor=red,citecolor=blue,urlcolor=blue]{hyperref}
\newcommand{\smCE}{\mathsf{smCE}}
\newcommand{\CalDist}{\mathsf{CalDist}}
\newcommand{\LowerCalDist}{\mathsf{LowerCalDist}}
\allowdisplaybreaks

\title{\huge A Matching Lower Bound for Online Smooth Calibration}
\author
{
   Kaixuan Ji\thanks{Department of Computer Science, University of California, Los Angeles, CA 90095, USA; e-mail: {\tt kaixuanji@cs.ucla.edu}} 
	~~~and~~~
	Quanquan Gu\thanks{Department of Computer Science, University of California, Los Angeles, CA 90095, USA; e-mail: {\tt qgu@cs.ucla.edu}}
}
\date{}

\begin{document}

\maketitle

\begin{abstract}
We study smooth calibration error in sequential binary prediction, a measure of how far probabilistic forecasts are from being calibrated that is robust to small perturbations of the forecasts. The minimax smooth calibration error is known to lie between $\Omega(T^{1/3})$ and $O(\sqrt T)$~\citep{qiao2024distance}, where $T$ is the number of rounds. We close this gap for both adaptive and oblivious adversaries. We construct an adaptive adversary whose outcome mean moves with the forecaster's cumulative residual, which makes compensating for past residuals costly and forces every forecaster to incur $\Omega(\sqrt T)$ expected smooth calibration error. Simulating this adversary against the forecaster's average prediction yields an oblivious adversary with the same guarantee. These lower bounds match the known $O(\sqrt T)$ upper bound up to universal constants, and are therefore minimax optimal. Our results also imply matching $\Theta(\sqrt T)$ bounds for calibration distance and lower calibration distance. The proof ideas are partially developed by GPT-6 Astra under the authors' guidance.
\end{abstract}

\section{Introduction}
We study the sequential binary forecasting problem of~\citet{foster1998asymptotic}. Over $T$ rounds, a forecaster and an adversary interact as follows. At each round $t \in [T]$, the forecaster predicts a probability $p_t \in [0,1]$ that the next binary outcome equals $1$, and the adversary simultaneously selects the outcome $x_t \in \{0,1\}$. Both $p_t$ and $x_t$ are then revealed, and may factor into the subsequent actions of both players. This problem abstracts many applications in which probabilistic predictions are issued sequentially, including weather forecasting~\citep{glenn1950verification,dawid1982well}, uncertainty quantification of machine learning models~\citep{guo2017calibration}, fair risk prediction~\citep{hebert2018multicalibration}, and forecasting for downstream decision makers~\citep{kleinberg2023ucalibration}.

A natural requirement for the predictions to be interpretable as probabilities is \emph{calibration}~\citep{dawid1982well}: among the rounds on which the forecaster predicts a value $\alpha$, the outcome should equal $1$ in an $\alpha$ fraction of them. Formally, the forecaster is perfectly calibrated if $\sum_{t=1}^T (x_t - p_t)\ind\{p_t = \alpha\} = 0$ for every $\alpha \in [0,1]$. The most widely used measure of the violation of this condition is the $\ell_1$ calibration error, also known as the expected calibration error (ECE),
\begin{align*}
    \mathrm{ECE}(x,p) = \sum_{\alpha \in [0,1]} \bigg|\sum_{t=1}^T (x_t - p_t)\ind\{p_t = \alpha\}\bigg|,
\end{align*}
where only the values $\alpha \in \{p_1, \ldots, p_T\}$ contribute to the sum. However, ECE is discontinuous in the predictions. Suppose that $x$ contains equally many $0$s and $1$s. Then predicting $p_t = 1/2$ in every round gives zero ECE. If each $p_t$ is instead perturbed to $1/2 + \epsilon_t$ for tiny, distinct, nonzero $\epsilon_t$, then every round falls into its own bin, and ECE jumps to $\Omega(T)$, although the predictions are essentially unchanged.

This sensitivity motivates the \emph{smooth calibration error}~\citep{kakade2008deterministic,blasiok2023unifying}, which tests the residuals against all bounded Lipschitz functions of the predictions,
\begin{align*}
    \smCE(x,p) = \sup_{f \in \cF}\sum_{t=1}^T f(p_t)(x_t - p_t),
\end{align*}
where $\cF$ is the class of $1$-Lipschitz functions from $[0,1]$ to $[-1,1]$. Since nearby predictions receive nearby weights, $\smCE$ is robust to small perturbations of the predictions, and in the example above it remains small.

For smooth calibration error, however, a gap remains in the sequential setting. \citet{qiao2024distance} gave a forecaster that achieves $O(\sqrt T)$ smooth calibration error against any adversary. It might appear obvious that this rate is tight, since independent fair outcomes make the cumulative residual $|\sum_t (x_t - p_t)|$ of order $\sqrt T$ when the forecaster predicts $1/2$. Surprisingly, \citet{qiao2024distance} showed that this argument is incorrect: against independent fair outcomes, a forecaster can compensate for past residuals with slightly biased predictions, and achieve only polylogarithmic smooth calibration error. Their adversary circumvents this by generating fair outcomes until the error is large and then stopping early with constant outcomes, which yields an $\Omega(T^{1/3})$ lower bound. This leaves the following question:
\begin{center}
\emph{Is $\Omega(\sqrt T)$ smooth calibration error unavoidable in sequential prediction?}
\end{center}

In this paper, we answer this question affirmatively, against both adaptive and oblivious adversaries. We first construct a new adaptive adversary whose outcome mean moves with the forecaster's cumulative residual, which makes compensating for past residuals costly, and show that it forces every forecaster to incur $\Omega(\sqrt T)$ expected smooth calibration error, improving the $\Omega(T^{1/3})$ lower bound of~\citet{qiao2024distance} and matching the known $O(\sqrt T)$ upper bound. We further convert this adversary into an oblivious one, which simulates the interaction against the forecaster's average prediction before the game, and show that the same lower bound holds even when the outcome sequence is fixed in advance.
Since smooth calibration error lower bounds both calibration distance and lower calibration distance up to a constant factor, our results also imply matching $\Theta(\sqrt T)$ bounds for these two measures in both settings.
A comparison of our results with directly related previous results is summarized in Table~\ref{tab:rates}.


\paragraph{Notation.}
For a positive integer $n$, let $[n]=\{1,\ldots,n\}$. We write $\EE$ and $\PP$ for expectation and probability, and $\ind\{\cdot\}$ for an indicator. We use calligraphic letters for sets and strategies. The notation $O(\cdot)$ and $\Omega(\cdot)$ hides universal positive constants, and $\Theta(\cdot)$ denotes matching bounds in both directions. We write $a \lesssim b$ if $a \leq Cb$ for a universal constant $C > 0$, and $a \asymp b$ if $a \lesssim b$ and $b \lesssim a$. 

\section{Related Work}\label{sec:related}

\paragraph{Sequential Calibration.}
Sequential calibration studies probabilistic forecasting of binary outcomes, where the forecaster issues a prediction before observing each outcome, and the outcomes may depend on previously revealed predictions. \citet{foster1998asymptotic} established the existence of randomized forecasters that achieve asymptotic calibration. \citet{foster1999proof} subsequently gave an algorithm based on Blackwell's approachability theorem. For the classical $\ell_1$ calibration error, these approaches yield an $O(T^{2/3})$ expected error, and \citet{hart2025calibrated} gave a direct minimax proof of this rate. \citet{dagan2025breaking} improved the upper bound to $O(T^{2/3-\epsilon})$ for a universal constant $\epsilon > 0$. For smooth calibration error and calibration distance, \citet{qiao2024distance} established an $O(\sqrt T)$ expected error. \citet{arunachaleswaran2025elementary} subsequently proposed the Almost-One-Step-Ahead algorithm, an explicit deterministic forecaster that guarantees calibration distance at most $2\sqrt T + 1$ on every outcome sequence. Calibration has also been studied across multiple subpopulations~\citep{hebert2018multicalibration,gupta2022online}, in omniprediction~\citep{garg2024oracle}, and for forecasts used by downstream decision makers~\citep{kleinberg2023ucalibration,hu2024predict}, including guarantees based on smooth calibration~\citep{hartline2025smooth,gopalan2026importance}.

\paragraph{Calibration Lower Bounds and Adversary Design.}
For the classical $\ell_1$ calibration error, independent fair outcomes already yield an $\Omega(\sqrt T)$ lower bound. \citet{qiao2021stronger} strengthened this to $\Omega(T^{0.528})$ via early stopping and sidestepping. \citet{dagan2025breaking} further obtained an $\Omega(T^{0.54389})$ lower bound with an oblivious adversary, whose outcomes are generated independently of the forecaster's predictions. For smooth calibration error and calibration distance, \citet{qiao2024distance} proved an $\Omega(T^{1/3})$ lower bound with an adversary that generates independent fair bits until a history-dependent stopping time and then repeats a fixed outcome. 
Our adversary retains the constant continuation after stopping, but replaces the fair bits with Bernoulli outcomes whose mean depends on the cumulative forecasting residual. 
The extension of this lower bound to oblivious adversaries follows the standard reduction from adaptive to oblivious adversaries in online learning~\citep{cesa2006prediction}, which does not apply to calibration error in general, since no deterministic forecaster achieves sublinear ECE~\citep{oakes1985self,dawid1985comment}.

\begin{table}[t]
\centering
\caption{Comparison of upper and lower bounds on the minimax expected calibration error in the simultaneous forecasting protocol. All of the upper bounds and our lower bounds hold against both adaptive and oblivious adversaries. In this table, $T\ge1$ is the total number of interactions, $\smCE$ stands for smooth calibration error, $\LowerCalDist$ stands for lower calibration distance, and $\CalDist$ stands for calibration distance. The lower bounds on lower calibration distance and calibration distance are direct consequences of the lower bound on smooth calibration error, as detailed in Remark~\ref{rmk:distance-based-calibration}.}
\label{tab:rates}
\vspace{1ex}
\begin{tabular}{lgggg}
\toprule
\rowcolor{white} Type & Reference & $\smCE$ & $\LowerCalDist$ & $\CalDist$ \\
\midrule
\rowcolor{white} Upper Bound & \citet{qiao2024distance} & $O(\sqrt T)$ & $O(\sqrt T)$ & $O(\sqrt T)$\\
\midrule
\rowcolor{white} & \citet{qiao2024distance} & $\Omega(T^{1/3})$ & $\Omega(T^{1/3})$ & $\Omega(T^{1/3})$ \\[3pt] 
\multirow{-2}{*}{Lower Bound} & This Work & $\Omega(\sqrt T)$ & $\Omega(\sqrt T)$ & $\Omega(\sqrt T)$ \rule{0pt}{2.5ex} \\
\bottomrule
\end{tabular}
\end{table}

\section{Preliminaries}\label{sec:prelim}
We consider an online binary forecasting game between a forecaster and an adversary over $T \geq 1$ rounds. At each round $t \in [T]$, the forecaster predicts a probability $p_t \in [0,1]$ and the adversary simultaneously selects an outcome $x_t \in \{0,1\}$, after which both are revealed. The prediction $p_t$ may be randomized and may depend on the history $(x_s, p_s)_{s<t}$, but not on the current outcome $x_t$. Formally, a forecaster strategy $\cA$ consists of an internal random variable $\xi$, drawn before the game independently of the adversary, and deterministic maps $(\cA_t)_{t=1}^T$, such that the prediction at round $t$ is $p_t = \cA_t(x_{<t}, p_{<t}; \xi)$, where $x_{<t} = (x_1, \ldots, x_{t-1})$ and $p_{<t} = (p_1, \ldots, p_{t-1})$; in particular, by induction, $p_t$ is a deterministic function of $x_{<t}$ and $\xi$. Therefore, for simplicity, we drop the dependence on the previous predictions and write $p_t = \cA_t(x_{<t};\xi)$ in place of $\cA_t(x_{<t}, p_{<t};\xi)$. Depending on how the adversary generates the revealed outcome $x_t$, we consider two settings.
\begin{itemize}[leftmargin=*]
\item In the \emph{adaptive} setting, $x_t$ may be randomized and may depend on the history $(x_s, p_s)_{s<t}$, but not on the current prediction $p_t$.
\item In the \emph{oblivious} setting, the adversary fixes the entire outcome sequence before the game, without observing the forecaster's predictions, and reveals one outcome per round. In particular, $x_t$ cannot depend on any of the forecaster's predictions.
\end{itemize}
We write $x \coloneqq (x_1,\ldots,x_T)$ and $p \coloneqq (p_1,\ldots,p_T)$ for the outcome and forecast sequences. The forecaster aims to keep the \emph{smooth calibration error}~\citep{kakade2008deterministic,blasiok2023unifying} of the realized sequences small. Our goal is to characterize the minimax smooth calibration error in both settings,
\begin{equation}\label{eq:minimax}
V_T^{\diamond} \coloneqq \inf_{\cA}\sup_{\cB \in \mathfrak{B}^{\diamond}}\EE_{\cA,\cB}[\smCE(x,p)], \qquad \diamond \in \{\mathrm{adp}, \mathrm{obv}\},
\end{equation}
where $\cA$ ranges over all forecaster strategies, $\mathfrak{B}^{\mathrm{adp}}$ and $\mathfrak{B}^{\mathrm{obv}}$ denote the sets of adaptive and oblivious adversary strategies, respectively, and $(x,p)$ is generated by the interaction between $\cA$ and $\cB$. That is, $\EE_{\cA,\cB}$ is taken over both the internal randomness $\xi$ of $\cA$ and the randomness of $\cB$. In particular, for a fixed outcome sequence $x \in \{0,1\}^T$, we write $\cB_x$ for the oblivious adversary that reveals $x$, in which case $\EE_{\cA,\cB_x}$ is taken over $\xi$ only. Since every oblivious adversary is a special case of an adaptive one, i.e., $\mathfrak{B}^{\mathrm{obv}} \subseteq \mathfrak{B}^{\mathrm{adp}}$, we have $V_T^{\mathrm{obv}} \leq V_T^{\mathrm{adp}}$. In both settings, $\smCE(x,p)$ is the smooth calibration error defined as follows.

\begin{definition}[Smooth calibration error]\label{def:smooth-calibration-error}
Let $\cF$ be the class of all $1$-Lipschitz functions from $[0,1]$ to $[-1,1]$. The smooth calibration error of $(x,p)$ is
\begin{equation}\label{eq:smdef}
\smCE(x,p)=\sup_{f\in\cF}\sum_{t=1}^T f(p_t)(x_t-p_t).
\end{equation}
\end{definition}

Definition~\ref{def:smooth-calibration-error} requires the forecaster to be calibrated around every value $u \in [0,1]$, while treating nearby forecasts as essentially the same. Specifically, for any fixed $u \in [0,1]$, among the rounds in which the forecaster predicts approximately $u$, the outcome should equal $1$ in approximately a $u$ fraction of them. Suppose this fails around some $u$, say the residuals $\sum_{t:p_t\approx u}(x_t-p_t)$ add up to a large positive amount. Then a test $f$ that is a small positive bump around $u$ and vanishes elsewhere collects this residual while ignoring all other rounds, so $\smCE$ is large. On the other hand, the Lipschitz constraint prevents $f$ from fluctuating rapidly near $u$. Consequently, the residuals $\sum_{t:p_t = u'}(x_t-p_t)$ for all $u' \approx u$ are effectively grouped together, and an over-prediction at one value can be offset by an under-prediction at a nearby value. This makes $\smCE$ robust to small perturbations of the forecasts, in contrast to the classical expected calibration error (ECE), which treats every forecast value as a separate bin and can change drastically under arbitrarily small perturbations. In short, $\smCE$ asks the forecaster to be calibrated at every scale coarser than the resolution of its forecasts, which is the property that matters when forecasts are used by downstream decision makers with smooth responses.

\begin{remark}[Distance-based measures]\label{rmk:distance-based-calibration}
Besides $\smCE$, two distance-based calibration measures have been studied: the \emph{calibration distance} and the \emph{lower calibration distance}~\citep{blasiok2023unifying,qiao2024distance}. The calibration distance is the $\ell_1$ distance from $p$ to the nearest perfectly calibrated forecast sequence,
\begin{align*}
\CalDist(x,p)=\inf_{z \in \cC(x)} \sum_{t=1}^T |p_t - z_t|,
\end{align*}
where $\cC(x)=\big\{ z \in[0,1]^T: \textstyle\sum_{t:z_t=u}(x_t-u)=0 \text{ for all } u\in[0,1]\big\}$ is the set of perfectly calibrated forecast sequences. The lower calibration distance relaxes this by allowing randomized repairs,
\begin{align*}
\LowerCalDist(x,p)=\inf_{\cD \in \underline{\cC}(x)} \sum_{t=1}^T \EE_{u \sim \cD_t}\big[|p_t - u|\big],
\end{align*}
where $\underline{\cC}(x)$ is the set of all $\cD = (\cD_t)_{t=1}^T$ such that each $\cD_t$ is a finitely supported distribution on $[0,1]$ and $\sum_{t=1}^T (x_t - \alpha)\cD_t(\alpha) = 0$ for every $\alpha \in [0,1]$. Since every $z \in \cC(x)$ corresponds to the point masses $\cD_t = \delta_{z_t}$ in $\underline{\cC}(x)$, we have $\LowerCalDist(x,p) \leq \CalDist(x,p)$. Moreover, \citet[Theorem~7.3]{blasiok2023unifying} implies $\frac12 \smCE(x,p) \leq \LowerCalDist(x,p)$. Therefore, any lower bound on $\smCE$ yields a lower bound of the same order on both $\LowerCalDist$ and $\CalDist$.
\end{remark}

\section{Adversaries and Lower Bounds}\label{sec:lower}

In this section, we show that the $O(\sqrt T)$ upper bound is tight against both adaptive and oblivious adversaries. In Section~\ref{sec:adaptive}, we construct a single adaptive adversary that forces every forecaster to incur $\Omega(\sqrt T)$ expected smooth calibration error. In Section~\ref{sec:oblivious}, we further extend this adversary, together with the induced lower bound, to oblivious adversaries.

\subsection{Adaptive Adversary}\label{sec:adaptive}
We summarize the adversary in Algorithm~\ref{alg:adv}, which follows the early-stopping design of~\citet{qiao2024distance}. In particular, the adversary keeps track of the cumulative residual
\begin{align*}
    S_t = \sum_{s=1}^t (x_s - p_s), \qquad S_0 = 0,
\end{align*}
and operates in two phases separated by the threshold $r = \sqrt T/3$: an \emph{active phase}, in which the outcomes are random, and a \emph{frozen phase}, in which the outcomes are constant. We explain the design of each phase in turn.

\paragraph{Freezing a Large Residual.} We start with the easy case. Note that $S_T$ is the score of the constant test $f \equiv 1$ and $-S_T$ is the score of $f \equiv -1$, so that $\smCE(x,p) \geq |S_T|$. Hence, once the residual reaches the threshold, i.e., $|S_t| \geq r$ at some round $t$, the adversary only needs to keep it there until the end. This is achieved by raising the flag $\mathit{frozen}$ and repeating the constant outcome $b = \ind\{S_t > 0\}$ in all remaining rounds. If $b = 1$, every later increment $x_t - p_t = 1 - p_t$ is nonnegative whatever the forecaster predicts, and symmetrically for $b = 0$. Therefore $|S_t|$ never decreases after the crossing, and $\smCE(x,p) \geq r = \Omega(\sqrt T)$.

\paragraph{Random Outcomes Force the Forecaster to Adjust.} It remains to handle forecasters that keep the residual below the threshold. In the active phase, the adversary draws $x_t \sim \operatorname{Bernoulli}(q_t)$ using fresh randomness, so that the residual behaves like a random walk. To see this, let $v_t = q_t(1-q_t)$ be the conditional variance of $x_t$. Ignoring the frozen phase for the moment and summing the expected increments of $S_t^2$ over all rounds, we obtain
\begin{align}\label{eq:square-main}
    \EE[S_T^2] \approx \EE \Bigg[ \sum_{t=1}^T v_t \Bigg] + 2\,\EE \Bigg[ \sum_{t=1}^T S_{t-1}(q_t - p_t) \Bigg].
\end{align}
When $q_t$ is bounded away from $0$ and $1$, $v_t$ is bounded away from zero, so the first sum is of order $T$. Hence, if the forecaster makes no adjustment, for example, $p_t = q_t$, then the residual grows to order $\sqrt T$ and reaches the threshold with constant probability. The only way to keep the residual small is to make the cross term in~\eqref{eq:square-main} negative, that is, to choose $p_t$ such that $q_t - p_t$ has the opposite sign of $S_{t-1}$, which pushes the residual back toward zero.

\paragraph{Penalizing the Adjustment.} Such adjustment is detected by the affine test $f(u) = 1/2 - u$. Indeed, conditioning on the history and the current prediction, we have
\begin{align}\label{eq:affine-increment-main}
    \EE\big[(1/2 - p_t)(x_t - p_t) \,\big|\, \text{history}, p_t\big] = (1/2 - p_t)(q_t - p_t) = (q_t - p_t)^2 + (1/2 - q_t)(q_t - p_t).
\end{align}
In~\citet{qiao2024distance}, the outcomes in the active phase are fair bits with $q_t = 1/2$, so the adjustment is penalized only by the quadratic term $(q_t - p_t)^2$. Being second order, this penalty is mild, and the forecaster can control the residual through many small adjustments at little cost. To make the penalty harsher, we let the mean depend on the residual,
\begin{equation}\label{eq:mean-main}
    q_t = \frac12 + \lambda S_{t-1}, \qquad \lambda = \frac{1}{4r},
\end{equation}
so that $1/2 - q_t = -\lambda S_{t-1}$ has the opposite sign of $S_{t-1}$. The second term in~\eqref{eq:affine-increment-main} then becomes $-\lambda S_{t-1}(q_t - p_t)$, which is positive precisely when the forecaster adjusts. In other words, the cross term that the forecaster must make negative to control the residual in~\eqref{eq:square-main} reappears with the opposite sign in the affine score. This intensified penalty is the key to the sharper lower bound.

\begin{algorithm}[t]
\caption{Residual-Dependent Adversary}\label{alg:adv}
\begin{algorithmic}[1]
\STATE \textbf{Input:} horizon $T$; set $r\gets\sqrt T/3$, $\lambda\gets1/(4r)$
\STATE $S\gets0$; $\mathit{frozen}\gets\mathsf{false}$
\FOR{$t=1,\ldots,T$}
\IF{$\mathit{frozen}$}
\STATE $x_t\gets b$
\ELSE
\STATE Draw $x_t\sim\operatorname{Bernoulli}(1/2+\lambda S)$
\ENDIF
\STATE Reveal $x_t$, observe the forecaster's prediction $p_t$, and set $S\gets S+x_t-p_t$
\IF{$\mathit{frozen}=\mathsf{false}$ and $|S|\ge r$}
\STATE $\mathit{frozen}\gets\mathsf{true}$; $b\gets\ind\{S>0\}$
\ENDIF
\ENDFOR
\end{algorithmic}
\end{algorithm}

\subsubsection{Lower Bounds under Adaptive Adversary}

The following theorem presents the lower bound induced by the adaptive adversary in Algorithm~\ref{alg:adv}. Its full proof is deferred to Appendix~\ref{app:proof-adaptive}.

\begin{theorem}[Lower bound against adaptive adversaries]\label{thm:main}
For every integer $T\ge1$, let $\cB_T$ denote the adversary in Algorithm~\ref{alg:adv}. Then every randomized forecasting strategy $\cA$ satisfies
\begin{align}\label{eq:mainlower}
\EE_{\cA,\cB_T}[\smCE(x,p)]\ge\frac{9}{155}\sqrt T.
\end{align}
Consequently, the minimax smooth calibration error against adaptive adversaries satisfies $V_T^{\mathrm{adp}} \geq 9\sqrt T/155$.
\end{theorem}

Theorem~\ref{thm:main} shows that the adversary in Algorithm~\ref{alg:adv} forces every forecaster to incur $\Omega(\sqrt T)$ expected smooth calibration error. This lower bound matches the $O(\sqrt{T})$ upper bounds of~\citet{qiao2024distance,arunachaleswaran2025elementary} up to a universal constant, and thus characterizes the minimax rate of smooth calibration error against adaptive adversaries.

\paragraph{Proof Overview.}We now give a brief overview of why this adversary achieves the $\Omega(\sqrt{T})$ lower bound. Adding $\lambda/2$ times~\eqref{eq:square-main} to the sum of~\eqref{eq:affine-increment-main} over all rounds cancels the cross terms and roughly leaves
\begin{align*}
    \EE \Bigg[\sum_{t=1}^T (1/2 - p_t)(x_t - p_t) \Bigg] + \frac\lambda2\EE[S_T^2] \geq \frac\lambda2 \EE \Bigg[\sum_{t=1}^T v_t \Bigg] \gtrsim \lambda T,
\end{align*}
whatever the forecaster does, where we drop the nonnegative terms involving $(q_t - p_t)^2$. 
Consequently, every forecaster faces a dichotomy: either the residual reaches the threshold and the frozen phase forces error at least $r$, or the game stays in the active phase for most of the $T$ rounds and the affine test collects a score of order $\lambda T \asymp T/r$. As a result, the forecaster always incurs an error of order at least $\min\{r, T/r\}$, and choosing $r \asymp \sqrt T$ yields the desired $\Omega(\sqrt{T})$ rate. 

\subsection{Oblivious Adversary}\label{sec:oblivious}

In this subsection, we extend the adaptive adversary of Section~\ref{sec:adaptive} to the oblivious setting. At a high level, the adversary simulates the entire interaction between Algorithm~\ref{alg:adv} and the forecaster before the game, with the forecaster's actual predictions replaced by their expectations. The detailed mechanism is summarized in Algorithm~\ref{alg:obl}, whose adversary we denote by $\cB^{\mathrm{obv}}_T$. In the following, we fix a forecasting strategy $\cA$ with internal random variable $\xi$, and recall from Section~\ref{sec:prelim} that its prediction at round $t$ is $\cA_t(x_{<t};\xi)$.

In particular, 
Algorithm~\ref{alg:obl} simulates the interaction between the adversary in Algorithm~\ref{alg:adv} and a deterministic forecaster $\bar\cA$, which predicts the average of $\cA$ over its internal randomness. Formally, given the past outcomes $x_{<t} = (x_1, \ldots, x_{t-1})$, $\bar\cA$ predicts
\begin{align}\label{eq:mean-prediction}
\mu_t = \EE_{\xi}\big[\cA_t(x_{<t};\xi)\big]
\end{align}
at each round $t$. Following Algorithm~\ref{alg:adv}, as long as the residual $S$ between the simulated outcomes and the predictions of $\bar\cA$ stays below the threshold $r$, the adversary draws $x_t \sim \operatorname{Bernoulli}(1/2 + \lambda S)$, whose mean moves with the residual. Once the threshold is reached, the adversary repeats the constant outcome that preserves the sign of the residual.
Since the simulation uses only the description of $\cA$ but not its realized predictions, the outcome sequence is independent of the internal randomness $\xi$, so that $\cB^{\mathrm{obv}}_T$ is a valid oblivious adversary.
After the simulation, the adversary reveals the simulated outcomes $x = (x_1, \ldots, x_T)$ one per round in the actual game, regardless of the forecaster's actual predictions.
Note that $\cB^{\mathrm{obv}}_T$ is a randomized oblivious adversary, i.e., a distribution over outcome sequences $x \in \{0,1\}^T$, rather than a single sequence. As we show below, replacing the forecaster's predictions with their expectations does not help the forecaster. Consequently, this distribution forces a large expected smooth calibration error on at least one sequence in its support, which yields the fixed sequence in Theorem~\ref{thm:oblivious}, so that the lower bound of Theorem~\ref{thm:main} carries over to oblivious adversaries.

\begin{algorithm}[t]
\caption{Oblivious Residual-Dependent Adversary}\label{alg:obl}
\begin{algorithmic}[1]
\STATE \textbf{Input:} horizon $T$ and forecasting strategy $\cA$; set $r\gets\sqrt T/3$, $\lambda\gets1/(4r)$
\STATE \textit{// Simulation before the game}
\STATE $S\gets0$; $\mathit{frozen}\gets\mathsf{false}$
\FOR{$t=1,\ldots,T$}
\STATE Set $x_t\gets b$ if $\mathit{frozen}$, and draw $x_t\sim\operatorname{Bernoulli}(1/2+\lambda S)$ otherwise
\STATE Set $S\gets S+x_t-\mu_t$, where $\mu_t \gets \EE_{\xi}[\cA_t(x_{<t};\xi)]$
\IF{$\mathit{frozen}=\mathsf{false}$ and $|S|\ge r$}
\STATE $\mathit{frozen}\gets\mathsf{true}$; $b\gets\ind\{S>0\}$
\ENDIF
\ENDFOR
\ENSURE $x = (x_1, \ldots, x_T)$
\end{algorithmic}
\end{algorithm}

\subsubsection{Lower Bounds under Oblivious Adversary}

The following theorem presents the lower bound induced by the oblivious adversary in Algorithm~\ref{alg:obl}. Its full proof is deferred to Appendix~\ref{app:proof-oblivious}.

\begin{theorem}[Lower bound against oblivious adversaries]\label{thm:oblivious}
For every integer $T \geq 1$ and every randomized forecasting strategy $\cA$, there exists an outcome sequence $x \in \{0,1\}^T$ such that the oblivious adversary $\cB_x$ satisfies
\begin{align}\label{eq:obl-lower}
\EE_{\cA,\cB_x}\big[\smCE(x, p)\big] \geq \frac{9}{155}\sqrt T,
\end{align}
where $p = (\cA_t(x_{<t};\xi))_{t=1}^T$ is the forecast sequence of $\cA$ on $x$.
Consequently, the minimax smooth calibration error against oblivious adversaries satisfies $V_T^{\mathrm{obv}} \geq 9\sqrt T/155$.
\end{theorem}

Theorem~\ref{thm:oblivious} shows that the $\Omega(\sqrt T)$ lower bound continues to hold against oblivious adversaries, with the same constant as Theorem~\ref{thm:main}. Since $V_T^{\mathrm{obv}} \leq V_T^{\mathrm{adp}}$, combining it with the $O(\sqrt T)$ upper bounds of~\citet{qiao2024distance,arunachaleswaran2025elementary} gives $V_T^{\mathrm{obv}} \asymp V_T^{\mathrm{adp}} \asymp \sqrt T$, which settles the minimax rate of smooth calibration error in both settings. 
Moreover, recall from Remark~\ref{rmk:distance-based-calibration} that $\smCE(x,p) \leq 2\LowerCalDist(x,p) \leq 2\CalDist(x,p)$ for every outcome sequence $x$ and forecast sequence $p$. Combining this with Theorem~\ref{thm:oblivious} immediately leads to the following corollary.

\begin{corollary}[Calibration distance lower bound]\label{cor:distance-lower}
For every integer $T\ge1$ and every randomized forecasting strategy $\cA$, the oblivious adversary $\cB_x$ in Theorem~\ref{thm:oblivious} satisfies
\begin{align*}
\EE_{\cA,\cB_x}\big[\CalDist(x, p)\big] \geq \EE_{\cA,\cB_x}\big[\LowerCalDist(x, p)\big] \geq \frac{9}{310}\sqrt T.
\end{align*}
\end{corollary}

Corollary~\ref{cor:distance-lower} matches, up to a universal constant, the $O(\sqrt T)$ upper bounds on both distance-based measures in~\citet{qiao2024distance}, which hold against adaptive and hence oblivious adversaries.

\paragraph{Proof Overview.} We now give a brief overview of why this adversary achieves the $\Omega(\sqrt T)$ lower bound. The proof of Theorem~\ref{thm:main} only uses the four tests $\cF_0 = \{u \mapsto \pm 1,\ u \mapsto 1-u,\ u \mapsto -u\}$, whose maximal score
\begin{align}\label{eq:phi}
\Phi(x,p) \coloneqq \max_{f \in \cF_0} \sum_{t=1}^T f(p_t)(x_t - p_t) = \max\Big\{|S(x,p)|,\ A(x,p) + \frac12|S(x,p)|\Big\}
\end{align}
is convex in $p$, since $S(x,p) = \sum_{t} (x_t - p_t)$ is affine and $A(x,p) = \sum_{t} (1/2 - p_t)(x_t - p_t)$ is a convex quadratic in $p$. Hence, for every fixed outcome sequence $x$, since $\Phi \leq \smCE$ and $\Phi(x, \cdot)$ is convex, Jensen's inequality gives $\EE_{\cA,\cB_x}[\smCE(x, p)] \geq \Phi(x, \EE_{\xi}[p])$, where $\EE_{\xi}[p] = (\EE_{\xi}[\cA_t(x_{<t};\xi)])_{t=1}^T$ is exactly the forecast sequence of $\bar\cA$ on $x$. That is, the forecaster $\cA$ performs no better than its deterministic average $\bar\cA$. Meanwhile, since Algorithm~\ref{alg:obl} simulates Algorithm~\ref{alg:adv} against $\bar\cA$, the proof of Theorem~\ref{thm:main} gives $\EE_{x \sim \cB^{\mathrm{obv}}_T}[\Phi(x, \mu)] \geq 9\sqrt T/155$, where $\mu = (\mu_1, \ldots, \mu_T)$ is the simulated forecast sequence. Combining the two bounds yields Theorem~\ref{thm:oblivious}.

\section{Conclusion}\label{sec:conclusion}
In this paper, we study the minimax smooth calibration error in sequential binary prediction against both adaptive and oblivious adversaries. In particular, we construct a new adaptive adversary whose outcome mean moves with the forecaster's cumulative residual and which switches to constant outcomes once the residual becomes large, so that any attempt to compensate for past residuals is penalized by an affine test. We further convert it into an oblivious adversary that simulates this interaction against the forecaster's average prediction before the game, which loses nothing since the tests used in our analysis have a convex maximal score. We show that both adversaries force every forecaster to incur $\Omega(\sqrt T)$ expected smooth calibration error, which improves the $\Omega(T^{1/3})$ lower bound of~\citet{qiao2024distance} and matches the known $O(\sqrt T)$ upper bound, thereby settling the minimax rates of smooth calibration error, as well as of calibration distance and its relaxation, in both settings up to universal constants.

\subsection*{Acknowledgment}

The proof ideas are partially developed by GPT-6 Astra under the authors’ guidance. The authors re-organized the proof, verified the proof
details, prepared the draft, and take full responsibility for the contents.

\appendix

\section{Proofs of Main Results}\label{sec:proof}

In this section, we provide the full proofs of Theorems~\ref{thm:main} and~\ref{thm:oblivious}. We first set up the notation and establish a lemma on the interaction with the adaptive adversary, which is shared by both proofs.

Fix an arbitrary, possibly randomized, forecasting strategy $\cA$, let $\xi$ be its internal random variable, and let $\cG_t = \sigma(\xi, x_1, \ldots, x_t)$ for $t = 0, 1, \ldots, T$. Since $p_t = \cA_t(x_{<t};\xi)$, the predictions $p_1, \ldots, p_t$ are $\cG_{t-1}$-measurable. Moreover, on the active rounds of Algorithm~\ref{alg:adv}, the outcome $x_t$ is drawn from $\operatorname{Bernoulli}(q_t)$ using fresh randomness that is independent of $\xi$ and of the history. We lower bound the calibration error through two kinds of tests, namely the constant tests $f \equiv 1$ and $f \equiv -1$, and the affine test $f(u) = 1/2 - u$, which lead to the following scores:
\begin{align*}
    S_t = \sum_{s=1}^{t}(x_s - p_s), \quad A_t = \sum_{s=1}^{t} (1/2 - p_s)(x_s - p_s).
\end{align*}
Moreover, we track the first round $\tau$ at which the residual reaches the threshold $r$,
\begin{align*}
    \tau = \min\big(\{t \in [T] : |S_t| \geq r\} \cup \{T\}\big),
\end{align*}
and let $H = \{|S_\tau| \geq r\}$ be the event that the threshold is reached, so that $\tau = T$ on $H^c$. Since $\max_{s \leq t} |S_s|$ is $\cG_t$-measurable, $\tau$ is a stopping time with respect to $\{\cG_t\}_{t=0}^T$.

\begin{lemma}\label{lem:key}
For every integer $T \geq 1$ and every randomized forecasting strategy $\cA$, let $\cB_T$ be the adaptive adversary in Algorithm~\ref{alg:adv}. Then the quantity $M \coloneqq \max\{|S_\tau|, A_\tau + |S_\tau|/2\}$ satisfies
\begin{align*}
\EE_{\cA,\cB_T}[M] \geq \frac{9}{155}\sqrt T.
\end{align*}
\end{lemma}

\begin{proof}
We first compute $\EE[A_\tau]$ as follows:
\begin{align*}
    \EE[A_\tau] &= \sum_{t=1}^T \EE\big[\ind\{t \leq \tau\}(1/2 - p_t)(x_t - p_t)\big] \\
    & = \sum_{t=1}^T \EE\Big[ \EE\big[\ind\{t \leq \tau\}(1/2 - p_t)(x_t - p_t) \big| \cG_{t-1} \big] \Big].
\end{align*}
For each fixed $t$, the indicator $\ind\{t \leq \tau\}$ is $\cG_{t-1}$-measurable, since $\{t \leq \tau\} = \{|S_s| < r \text{ for all } s < t\}$, and so are $p_t$ and $q_t = 1/2 + \lambda S_{t-1}$. Moreover, on $\{t \leq \tau\}$, round $t$ is active, so that $\EE[x_t \mid \cG_{t-1}] = q_t$. Therefore,
\begin{align*}
    \EE\big[\ind\{t \leq \tau\}(1/2 - p_t)(x_t - p_t) \big| \cG_{t-1} \big] & = \ind\{t \leq \tau\}(1/2 - p_t)(q_t - p_t) \\
    & = \ind\{t \leq \tau\}\big[(q_t - p_t)^2 - \lambda S_{t-1}(q_t - p_t)\big],
\end{align*}
where the last equality uses $1/2 - p_t = (1/2 - q_t) + (q_t - p_t)$ and $1/2 - q_t = -\lambda S_{t-1}$. Consequently, we see that
\begin{align}
    \EE[A_\tau] & = \sum_{t=1}^T \EE\Big[\ind\{t \leq \tau\}\big[(q_t - p_t)^2 - \lambda S_{t-1}(q_t - p_t)\big]\Big] \nonumber \\
    & = \EE \Bigg[ \sum_{t=1}^{\tau}\Big[(q_t - p_t)^2 - \lambda S_{t-1}(q_t - p_t)\Big] \Bigg]. \label{eq:EA-main}
\end{align}
Similarly, we write $S_\tau^2$ as the telescoping sum $S_\tau^2 = \sum_{t=1}^T \ind\{t \leq \tau\}(S_t^2 - S_{t-1}^2)$, where each increment satisfies $S_t^2 - S_{t-1}^2 = 2S_{t-1}(x_t - p_t) + (x_t - p_t)^2$. Therefore,
\begin{align*}
    \EE[S_\tau^2] & = \sum_{t=1}^T \EE\Big[\ind\{t \leq \tau\}\big(2S_{t-1}(x_t - p_t) + (x_t - p_t)^2\big)\Big] \\
    & = \sum_{t=1}^T \EE\Big[\EE\big[\ind\{t \leq \tau\}\big(2S_{t-1}(x_t - p_t) + (x_t - p_t)^2\big) \,\big|\, \cG_{t-1}\big]\Big].
\end{align*}
For each fixed $t$, all of $\ind\{t \leq \tau\}$, $S_{t-1}$, $p_t$, and $q_t$ are $\cG_{t-1}$-measurable. Moreover, on $\{t \leq \tau\}$, we have $\EE[x_t - q_t \mid \cG_{t-1}] = 0$ and $\EE[(x_t - q_t)^2 \mid \cG_{t-1}] = v_t$, where $v_t = q_t(1 - q_t)$ is the variance of $\operatorname{Bernoulli}(q_t)$. Therefore, writing $x_t - p_t = (x_t - q_t) + (q_t - p_t)$ and expanding,
\begin{align*}
    \EE\big[\ind\{t \leq \tau\}\big(2S_{t-1}(x_t - p_t) + (x_t - p_t)^2\big) \,\big|\, \cG_{t-1}\big] = \ind\{t \leq \tau\}\big[v_t + (q_t - p_t)^2 + 2S_{t-1}(q_t - p_t)\big],
\end{align*}
since all terms that are linear in $x_t - q_t$ vanish in conditional expectation. Consequently, we see that
\begin{align}
    \EE[S_\tau^2] & = \sum_{t=1}^T \EE\Big[\ind\{t \leq \tau\}\big[v_t + (q_t - p_t)^2 + 2S_{t-1}(q_t - p_t)\big]\Big] \nonumber \\
    & = \EE\Bigg[\sum_{t=1}^{\tau}\Big[v_t + (q_t - p_t)^2 + 2S_{t-1}(q_t - p_t)\Big]\Bigg]. \label{eq:ES-main}
\end{align}
Adding $\lambda/2$ times~\eqref{eq:ES-main} to~\eqref{eq:EA-main} cancels the cross terms, which yields
\begin{align}\label{eq:identity-main}
    \EE[A_\tau] + \frac\lambda2 \EE[S_\tau^2] = \Big(1 + \frac\lambda2\Big)\EE \Bigg[\sum_{t=1}^{\tau}(q_t - p_t)^2 \Bigg] + \frac\lambda2 \EE\Bigg[\sum_{t=1}^{\tau} v_t \Bigg].
\end{align}

Now, taking expectations in the bound $M \geq A_\tau + |S_\tau|/2$ and applying~\eqref{eq:identity-main}, we obtain
\begin{align*}
    \EE[M] \geq \EE[A_\tau] + \frac12\EE\big[|S_\tau|\big] \geq \frac\lambda2 \EE\Bigg[\sum_{t=1}^{\tau} v_t\Bigg] + \frac12 \EE\big[|S_\tau| - \lambda S_\tau^2\big],
\end{align*}
where we drop the nonnegative term $(1 + \lambda/2)\EE\big[\sum_{t \leq \tau}(q_t - p_t)^2\big]$. We now show that both remaining terms are well behaved. First, since $|S_{\tau-1}| < r$ and each increment of the residual has magnitude at most one, we have $|S_\tau| \leq r + 1$. Since $r = \sqrt T/3 \geq 1/3$, we further have $r + 1 \leq 4r = 1/\lambda$, so that $\lambda |S_\tau| \leq 1$ and the last term is nonnegative. Second, on every active round $t \leq \tau$, we have $|S_{t-1}| < r$, so that $|q_t - 1/2| = \lambda|S_{t-1}| < \lambda r = 1/4$ by the choice $\lambda = 1/(4r)$. Hence $q_t \in (1/4, 3/4)$, which implies $v_t \geq 3/16$. Therefore, writing $m = \EE[M]$, we obtain
\begin{align}\label{eq:time-main}
    m \geq \frac{3\lambda}{32}\EE[\tau] = \frac{9}{128\sqrt T}\EE[\tau].
\end{align}
That is, a long active phase forces a large affine score. On the other hand, we have $M \geq |S_\tau| \geq r\,\ind\{H\}$, so that $\PP(H) \leq m/r = 3m/\sqrt T$, i.e., reaching the threshold often forces a large constant score. Since $\tau = T$ on $H^c$, it follows that
\begin{align*}
    \EE[\tau] \geq T\,\PP(H^c) \geq T - 3m\sqrt T.
\end{align*}
Substituting this into~\eqref{eq:time-main} gives $m \geq 9\sqrt T/128 - 27m/128$, i.e., $m \geq 9\sqrt T/155$. This completes the proof of Lemma~\ref{lem:key}.
\end{proof}

\subsection{Proof of Theorem~\ref{thm:main}}\label{app:proof-adaptive}

\begin{proof}[Proof of Theorem~\ref{thm:main}]
By Lemma~\ref{lem:key}, it suffices to show that $\smCE(x,p)$ is lower bounded by $M = \max\{|S_\tau|, A_\tau + |S_\tau|/2\}$. If $\tau < T$, the threshold is reached at round $\tau$, and every later outcome equals $b = \ind\{S_\tau > 0\}$. Hence every later increment $x_t - p_t$ has the same sign as $S_\tau$. If $\tau = T$, there are no later rounds. Therefore, the constant test $f \equiv 1$ if $S_\tau \geq 0$, and $f \equiv -1$ otherwise, has score $|S_T| \geq |S_\tau|$, which gives the first lower bound. For the second lower bound, suppose $S_\tau \geq 0$ and consider $f(u) = 1 - u = (1/2 - u) + 1/2$. Its score over the first $\tau$ rounds is $A_\tau + S_\tau/2$, and its score in every later round is $(1 - p_t)(1 - p_t) \geq 0$. Similarly, if $S_\tau < 0$, the test $f(u) = -u = (1/2 - u) - 1/2$ has score $A_\tau - S_\tau/2$ over the first $\tau$ rounds and score $(-p_t)(0 - p_t) \geq 0$ in every later round. All these tests belong to the class $\cF_0 \subseteq \cF$ of the four tests in~\eqref{eq:phi}. In summary, we have
\begin{align}\label{eq:preserve-main}
    \smCE(x,p) \geq \Phi(x,p) \geq \max \Big\{|S_\tau|, A_\tau + \frac12 |S_\tau|\Big\} = M.
\end{align}
Taking expectations and applying Lemma~\ref{lem:key} completes the proof of Theorem~\ref{thm:main}.
\end{proof}

\subsection{Proof of Theorem~\ref{thm:oblivious}}\label{app:proof-oblivious}

We first show that, for a fixed outcome sequence, randomizing the predictions cannot decrease the restricted objective $\Phi$ in~\eqref{eq:phi} below its value at the averaged predictions.

\begin{lemma}[Convexity]\label{lem:convex}
For every fixed $x \in \{0,1\}^T$, the map $p \mapsto \Phi(x,p)$ is convex on $[0,1]^T$. Consequently, every random vector $y \in [0,1]^T$ satisfies
\begin{align}\label{eq:jensen}
\EE[\smCE(x,y)] \geq \EE[\Phi(x,y)] \geq \Phi(x,\EE[y]).
\end{align}
\end{lemma}

\begin{proof}
The score $S(x,p) = \sum_{t=1}^T (x_t - p_t)$ is affine in $p$, so $|S(x,p)|$ is convex in $p$. Moreover, expanding each summand of $A(x,p)$ gives
\begin{align*}
A(x,p) = \sum_{t=1}^T (1/2 - p_t)(x_t - p_t) = \sum_{t=1}^T \Big(p_t^2 - \big(x_t + 1/2\big)p_t + \frac{x_t}{2}\Big),
\end{align*}
which is a sum of convex quadratic functions of $p$. Hence $A(x,p) + |S(x,p)|/2$ is convex in $p$. Since the pointwise maximum of convex functions is convex, $\Phi(x,\cdot)$ is convex by~\eqref{eq:phi}. The first inequality in~\eqref{eq:jensen} follows from $\Phi \leq \smCE$, since $\cF_0 \subseteq \cF$, and the second inequality follows from Jensen's inequality.
\end{proof}

We are now ready to prove Theorem~\ref{thm:oblivious}.

\begin{proof}[Proof of Theorem~\ref{thm:oblivious}]
Fix a randomized forecasting strategy $\cA$ with internal random variable $\xi$, and let $\bar\cA$ be the deterministic forecaster in Algorithm~\ref{alg:obl}, which predicts $\EE_{\xi}[\cA_t(x_{<t};\xi)]$ at each round $t$. For every $x \in \{0,1\}^T$, define the expected smooth calibration error of $\cA$ on $x$ and the forecast sequence of $\bar\cA$ on $x$, respectively, as
\begin{align*}
g(x) \coloneqq \EE_{\cA,\cB_x}\big[\smCE(x, p)\big], \qquad \bar p(x) \coloneqq \big(\EE_{\xi}[\cA_t(x_{<t};\xi)]\big)_{t=1}^T,
\end{align*}
where $p = (\cA_t(x_{<t};\xi))_{t=1}^T$ is the forecast sequence of $\cA$ on $x$. For every fixed $x$, the outcome sequence does not depend on $\xi$, so that $\EE_{\xi}[p] = \bar p(x)$ by definition. Applying Lemma~\ref{lem:convex} with $y = p$, we obtain
\begin{align}\label{eq:obl-pointwise}
g(x) \geq \Phi\big(x, \bar p(x)\big) \quad \text{for every } x \in \{0,1\}^T.
\end{align}

Next, let $x = (x_1, \ldots, x_T)$ be the random outcome sequence generated by $\cB^{\mathrm{obv}}_T$ in Algorithm~\ref{alg:obl}. By construction, the simulation in Algorithm~\ref{alg:obl} is exactly the interaction between Algorithm~\ref{alg:adv} and the deterministic forecaster $\bar\cA$, whose prediction at round $t$ is $\bar p(x)_t = \EE_{\xi}[\cA_t(x_{<t};\xi)]$. Consequently, $(x, \bar p(x))$ has the same distribution as the outcome and forecast sequences generated by the interaction between $\bar\cA$ and $\cB_T$. Applying~\eqref{eq:preserve-main} and Lemma~\ref{lem:key} to the deterministic forecaster $\bar\cA$, we obtain
\begin{align}\label{eq:obl-sim}
\EE_{x \sim \cB^{\mathrm{obv}}_T}\Big[\Phi\big(x, \bar p(x)\big)\Big] = \EE_{\bar\cA,\cB_T}\big[\Phi(x,p)\big] \geq \frac{9}{155}\sqrt T.
\end{align}
Since the left-hand side of~\eqref{eq:obl-sim} is an average of $\Phi(x, \bar p(x))$ over $x \sim \cB^{\mathrm{obv}}_T$, there exists $x \in \{0,1\}^T$ such that $\Phi(x,\bar p(x)) \geq 9\sqrt T/155$. Plugging this into~\eqref{eq:obl-pointwise}, we obtain that, for this specific $x$,
\begin{align*}
    \EE_{\cA,\cB_x}\big[\smCE(x, p)\big] \geq \Phi\big(x,\bar p(x)\big) \geq \frac{9}{155}\sqrt T,
\end{align*}
which proves~\eqref{eq:obl-lower}. Finally, since $\cB_x \in \mathfrak{B}^{\mathrm{obv}}$ and $\cA$ is arbitrary,~\eqref{eq:minimax} gives $V_T^{\mathrm{obv}} \geq 9\sqrt T/155$, which completes the proof of Theorem~\ref{thm:oblivious}.
\end{proof}

\bibliographystyle{ims}
\bibliography{reference}

\end{document}